\documentclass[10pt,twocolumn,letterpaper]{article}

\usepackage[pagenumbers]{iccv} 
\usepackage{multirow} 
\usepackage{arydshln} 

\usepackage{tabularx}
\usepackage{booktabs}
\usepackage{dashrule}

\usepackage{multirow}
\usepackage{amsmath}

\definecolor{iccvblue}{rgb}{0.21,0.49,0.74}
\usepackage[pagebackref,breaklinks,colorlinks,allcolors=iccvblue]{hyperref}

\def\paperID{14077} 
\def\confName{ICCV}
\def\confYear{2025}

\title{Efficient Listener: Dyadic Facial Motion Synthesis via Action Diffusion}

\author{Zesheng Wang\\
Nantes Université\\
\and
Alexandre Bruckert\\
Nantes Université\\
\and
Patrick Le Callet\\
Nantes Université\\
\and
Guangtao Zhai\\
Shanghai Jiao Tong University\\
}

\begin{document}
\twocolumn[{
\maketitle
\begin{center}
    \captionsetup{type=figure}
    \includegraphics[width=\textwidth]{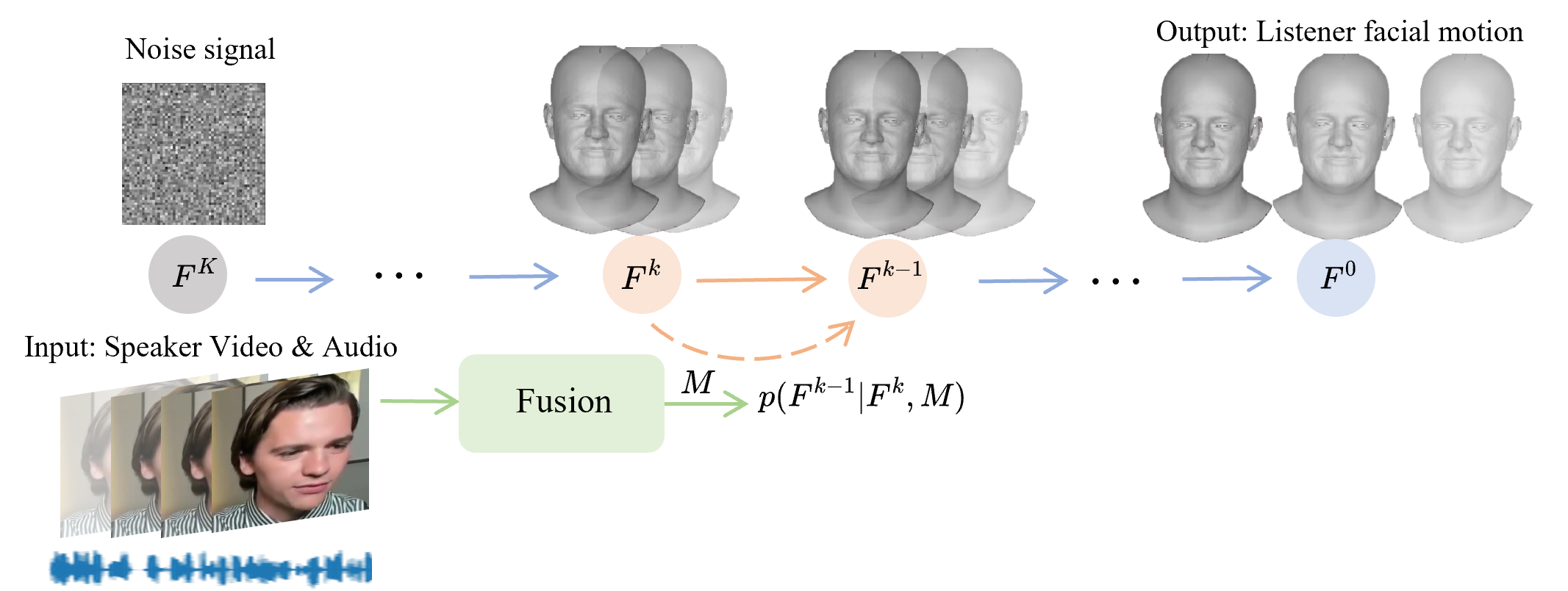}
    \captionof{figure}{Facial Action Diffusion in the forward process. We treat the visual and auditory perception of the speaker as conditional probability parameters to guide the listener's facial motion synthesis through a denoising process. Although our initial setting is not a Gaussian noise image, we use it as a metaphorical description of our noise signal.}
  \label{fig:fig1}
\end{center}
}]
\maketitle

\begin{abstract}
Generating realistic listener facial motions in dyadic conversations remains challenging due to the high-dimensional action space and temporal dependency requirements. Existing approaches usually consider extracting 3D Morphable Model (3DMM) coefficients and modeling in the 3DMM space. However, this makes the computational speed of the 3DMM a bottleneck, making it difficult to achieve real-time interactive responses. 
To tackle this problem, we propose Facial Action Diffusion (FAD), which introduces the diffusion methods from the field of image generation to achieve efficient facial action generation. We further build the Efficient Listener Network (ELNet) specially designed to accommodate both the visual and audio information of the speaker as input.
Considering of FAD and ELNet, the proposed method learns effective listener facial motion representations and leads to improvements of performance over the state-of-the-art methods while reducing 99\% computational time. 
\end{abstract}
    
\section{Introduction}
\label{sec:intro}
In interpersonal interactions, facial behavior plays a significant role as a non-verbal signal \cite{MAVRIDIS201522}. During face-to-face communication, the roles of speaker and listener are constantly interchanged as the interaction progresses. The speaker typically obtains feedback by observing the facial cues of the listener\cite{jack2015human}. This real-time, subtle feedback and interaction maintain the dynamic nature of the conversation \cite{tran2024dim}. This is particularly evident in online video call scenarios, where latency and packet loss significantly impact the communication experience \cite{BINGOL2024110356}. Therefore, the time constraint when generating facial motions of listeners is a particularly significant aspect of the task.

Unlike the task of speaker motion generation \cite{gan2023efficient, zhang2023sadtalker}, which typically uses audio \cite{wang2022one} or text \cite{li2021write} as driving signals to generate appropriate facial movements, the listener has to simultaneously process both visual and auditory information from the speaker. In particular, visual perception becomes a barrier to the computational efficiency of the generative models. Recently, some researchers have attempted to extract the speaker coefficients from the speaker video using 3D Morphable Models (3DMM), thereby enabling the modeling of the high-dimensional facial motion space \cite{ng2022learning, song2023emotional, zhou2022responsive}. Specifically, Zhou et al. \cite{zhou2022responsive} employed LSTM \cite{graves2012long} networks to learn the mapping from the speaker 3DMM coefficients and audio features to the listener facial motion responses. Ng et al. \cite{ng2022learning} introduced the VQ-VAE \cite{van2017neural} and subsequently learned the motion categories within the codebook to generate realistic listener facial motion. Song et al. \cite{song2023emotional} expanded the codebook space and incorporated emotion perception. However, the performance bottleneck of these methods is limited by the 3DMM. Despite Luo et al. \cite{luo2024reactface} employed visual perception via Transformer to improve inference speed, the performance remains constrained by the model architecture. 

Overall, the listener facial behavior is modeled as a transformation from visual and auditory perception space to a high-dimensional motion space, which cannot be naively aligned directly through feature dimensions. Moreover, interaction is not merely about listening. When the roles in an interaction switch rapidly, it is necessary to generate listener feedback flexibly and swiftly to minimize the impact of latency on communication. To address the aforementioned pressing issues, we focus on the following two key aspects: (1) How to model the transformation from visual and auditory perception to the high-dimensional facial motions of the listener; (2) How to generate the listener facial feedback quickly and flexibly.

To better model visual and auditory perception into high-dimensional dynamic facial motion, we propose Facial Action Diffusion (FAD). As shown in Figure \ref{fig:fig1}, we regard the generation process as a "conditional denoising diffusion process \cite{ho2020denoising}", which enables the proposed method generating facial motion efficiently and elegantly. First, FAD does not require the prior extraction of the speaker facial motion information (\eg facial expression, head rotation, \etc), directly utilizing the speaker visual and auditory perception. Second, FAD does not attempt to align visual and auditory multimodal perceptual information through feature dimensions. We incorporate visual and auditory information as conditional input in the diffusion process. Therefore, we are not required to design complex models and can opt for more lightweight models to enhance computational efficiency. 

However, establishing an efficient generative network using the proposed FAD approach is non-trivial. The network should adhere to two principles: (1)  It should dynamically adjust its computational strategy according to the duration of the speaker input segment, (2) It should be lightweight to enhance computational efficiency. Based on these two principles, we propose Efficient Listener Network (ELNet). First, to address the issue of varying durations of speaker input segments, we define the \textbf{clip}, the smallest processing unit of the input segment. Specifically, we divide the input video and audio into multiple \textbf{clips} using frame alignment. Subsequently, we input these \textbf{clips} into our framework in batches simultaneously. Moreover, we employ the U- \cite{ronneberger2015u} architecture, which has proven successful in image-based diffusion models, as the backbone and replace 2D convolutions with 1D convolutions. We evaluate our method on two popular datasets. Our contributions and innovations are summarized as follows.

\begin{itemize}[leftmargin=2em]
\item[\textbullet] We propose a novel paradigm for facial motion synthesis, FAD, which can efficiently and elegantly generate listener facial motions using visual and auditory perception.
\item[\textbullet] We introduce ELNet, which utilizes a lightweight model. It can efficiently process speaker input segments of diverse durations at the granularity of \textbf{clips} as the smallest unit.
\item[\textbullet] Extensive experiments demonstrate that our method not only maintains a leading position in qualitative results but also reduces computational time by 99\% compared to state-of-the-art methods.
\end{itemize}

\section{Related Work}
\label{sec:formatting}
Previous work has achieved excellent results in modeling listener facial motions. However, there are still some challenges that remaining, such as real-time human-computer interaction motion generation.
\paragraph{Facial Motion Generation for Listener.} 
In human-computer interaction, especially in dyadic interactions, one party typically acts as the speaker while the other acts as the listener. Methods for generating the speaker behavior have long received widespread attention \cite{cheng2022videoretalking, prajwal2020lip, kr2019towards}. Most methods have achieved success in modeling facial behavior from audio \cite{chen2019hierarchical, zhou2019talking}. However, in contrast to the task of generating speaker facial motions, the role of the listener faces greater challenges. As a listener, one needs to provide appropriate feedback in response to both the visual and auditory perceptions of the speaker \cite{greenwood2017predicting}. Moreover, humans are highly sensitive to natural human motion \cite{mori2012uncanny}. Early listener models were mostly rule-based, speech-driven interactive agents \cite{watanabe2004interactor}. Nishimura et al. \cite{nishimura2007spoken} proposed a decision tree model driven by prosodic audio features. Morency et al. \cite{morency2008predicting} demonstrated the feasibility of data-driven approaches using Hidden Markov Models (HMMs) and Conditional Random Fields (CRFs). In recent years, with the advancement of deep learning, data-driven motion generation methods have shown more promising development potential. Several researchers have focused on modeling listener facial motions. Song et al. \cite{song2023react2023} combined three dyadic interaction datasets collected under laboratory conditions: the NOXi \cite{cafaro2017noxi}, UDIVA \cite{palmero2021context}, and RECOLA \cite{ringeval2013introducing} datasets, to establish a listener facial motion generation dataset. To investigate the task of listener facial motion prediction in real-world interactions, Ng et al. \cite{ng2022learning} collected data from online dyadic interactions on YouTube. Zhou et al. \cite{zhou2019talking} collected diverse dyadic interaction data from the internet to build a dataset. The dataset not only contains video information but also includes meticulous manual annotations. Although existing methods have achieved some progress, there are still many challenges in predicting listener motions. To endow listeners with appropriate emotions, Song et al. \cite{song2023emotional} proposed the Emotional Listener Portrait. To obtain more diverse listener facial motions, Luo et al. \cite{luo2024reactface} developed ReactFace. However, to achieve real-time human-computer interaction, current methods still face difficulties in rapidly and efficiently generating appropriate listener facial motions.

\paragraph{Conditional Generative Models.} 
In dyadic interactions, listeners typically provide feedback after receiving visual and auditory stimuli from the speaker. This is essentially a conditional modeling process. Conditional networks are defined as generating corresponding outputs based on conditions (such as music \cite{li2021dance}, speech \cite{fan2022faceformer}, text \cite{zhang2024motiondiffuse}, and actions \cite{guo2020action2motion}). 
In the fields of gesture generation and human motion synthesis, these methods have continuously achieved better performance with the advancement of generative models. For example, Generative Adversarial Networks (GANs) \cite{barsoum2018hp} and Variational Autoencoders (VAEs) \cite{petrovich2021action} have shown promising results. Existing listener facial motion generation networks include Generative Adversarial Networks (GANs), which involve designing in an adversarial manner and are typically challenging to train \cite{barsoum2018hp}. Autoencoders (AEs) \cite{bank2023autoencoders} and Variational Autoencoders (VAEs) \cite{tomczak2018vae} are currently the most widely used methods for listener facial motion generation. Previous works, such as L2L \cite{ng2022learning}, introduced VQ-VAE \cite{van2017neural} to learn the embedding of motion sequences into a codebook. The ELP \cite{song2023emotional} expanded the codebook in terms of emotions. 
Recently, the remarkable performance of diffusion models in the field of image generation \cite{ramesh2022hierarchical} has attracted widespread attention. 
Given the excellent controllability of diffusion methods, human motion generation models inspired by diffusion models hold great potential \cite{zhi2023livelyspeaker}. We believe that the mapping capability of diffusion methods across modalities is well-suited for addressing the task of listener facial motion prediction.

\paragraph{Diffusion-based Motion Generation.}
Inspired by the success of diffusion models in text-to-image generation, Zhang et al. \cite{zhang2024motiondiffuse} designed a text-driven framework for human motion generation. Tseng et al. \cite{tseng2023edge} designed a transformer-based diffusion method to generate human motions from music. Zhi et al. \cite{zhi2023livelyspeaker} used a diffusion model to generate rhythmic information from audio. Motivated by these successful applications, we consider leveraging the unique characteristics of diffusion models, in combination with the specific requirements of listener facial motion generation, to introduce diffusion models for processing the visual and auditory perceptions obtained by the listener.
\section{Method}

\begin{figure*}[!t]\centering
    \captionsetup{type=figure}
    \includegraphics[width=\textwidth]{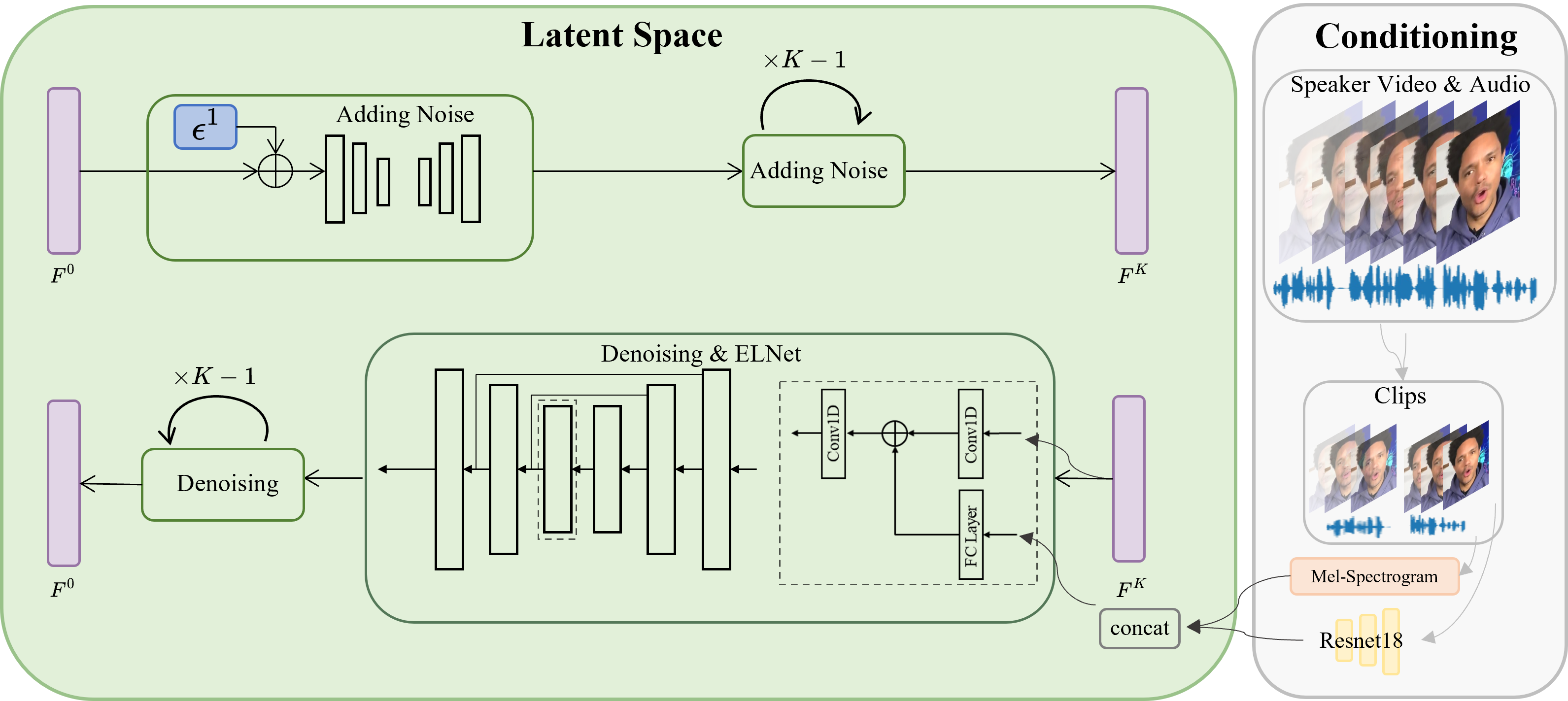}
    \captionof{figure}{The overview of the proposed method. We use multimodal data, including visual and auditory perception, as input to predict the listener facial feedback. We process the input at the granularity of \textbf{clips} as the smallest unit to achieve efficient handling of videos of varying lengths. Simultaneously, we innovatively regard the multimodal data as conditional probabilities for the diffusion noise addition and denoising processes, guiding the appropriate listener facial motions.
    }
  \label{fig:fig2}
\end{figure*}
\subsection{Overview}
Figure \ref{fig:fig2} illustrates an overview of our proposed method.
Given the speaker video $V$ and audio $A$ as inputs, our model can rapidly generate the corresponding listener facial motions $F$ at each moment. The listener facial motion is represented by 3DMM coefficients. It encompasses both the listener facial expressions and head movements $F=\{\beta ,R\}$.

The proposed framework can be divided into three parts: (1) Visual and Auditory Perception (Section \ref{sec:vap}). At this stage, we define the \textbf{Clip} as the smallest processing unit. Clips are fed into the model in batches to achieve high efficiency. To model the speaker visual and audio inputs, we introduce a lightweight convolutional network and integrate them with frame alignment. (2) Facial Action Diffusion (Section \ref{sec:fad}). 
To model the representation of listener facial motion from multimodal inputs, we introduce facial action diffusion, which maps the input features to a high-dimensional motion space through the processes of noise addition and denoising. (3) Efficient Listener Network (Section \ref{sec:elnet}). To learn motion representations within the diffusion process, we introduce Efficient Listener Network, a 1D convolutional network based on the U-Net architecture. The lightweight network effectively learns the feature representations in the latent space while maintaining high efficiency.

\subsection{Visual and Auditory Perception}
\label{sec:vap}
\begin{figure}[t]
  \centering
   \includegraphics[width=\linewidth]{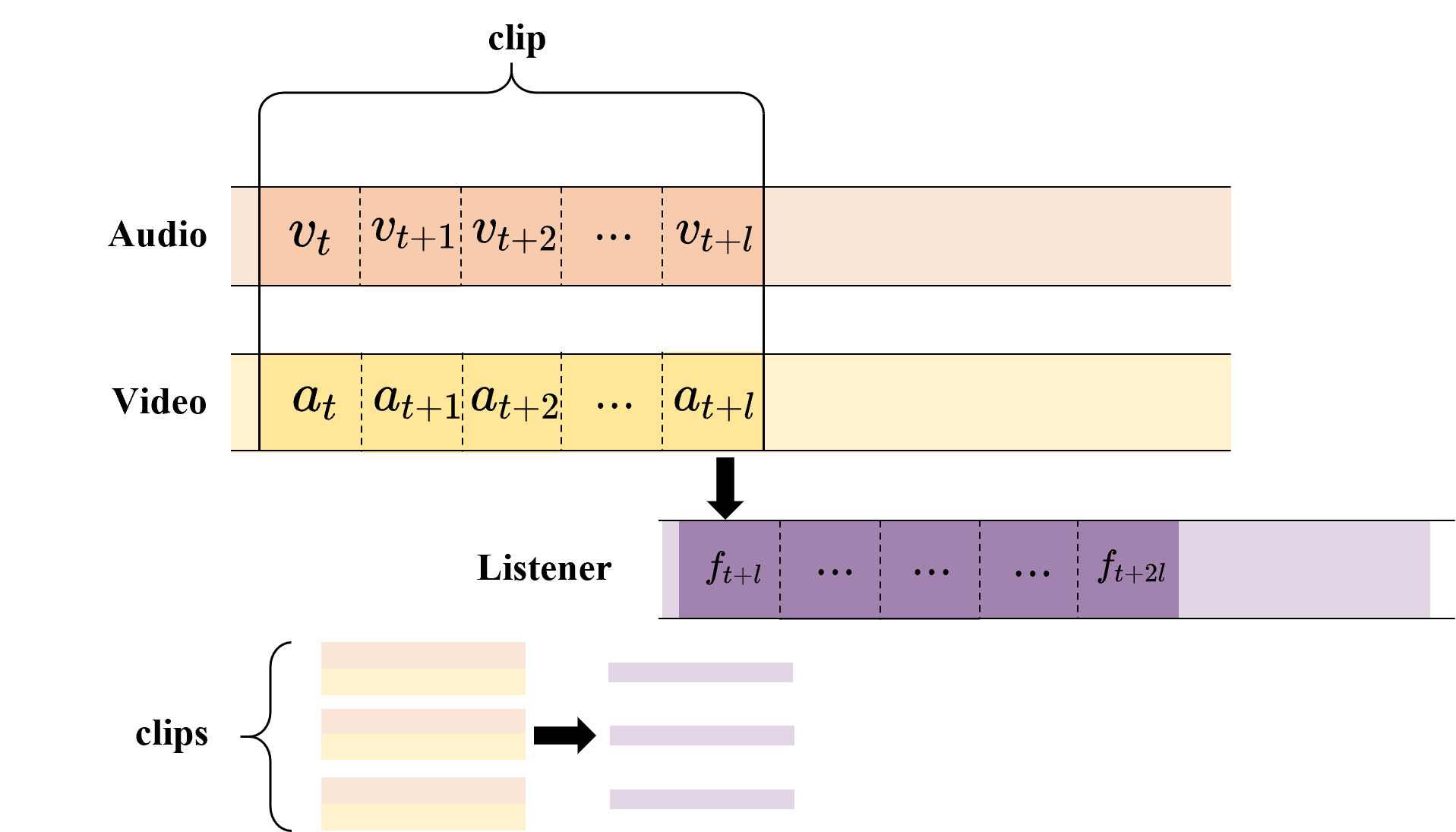}
   \caption{The input video and audio are processed in \textbf{clips}. We predict the future listener facial motions using batch processing.}
   \label{fig:fig3}
\end{figure}

We define $V_{T}=\{v_t\}_{t=1}^{T}$ and $A_{T}=\{a_t\}_{t=1}^{T}$ as the sequence of input video and audio with the length of $T$. Then, we define the \textbf{clip}, which enables efficient and flexible handling of inputs of varying lengths.
Specifically, we segment $V_{T}$ and $A_{T}$ along the temporal dimension into $n$ \textbf{clips} of minimum unit length \( l \) in a frame-aligned manner and represent them as $C_{i}=\{v_{i:(i+1)\times l},a_{i:(i+1)\times l}\}_{i=0}^{n-1}$. Based on the \textbf{clips}, we can generate the future listener facial motions $F_{i} = \{f_{i\times l:(i+1)+2l}\}_{i=0}^{n-1}$ using batch processing to improve computational efficiency. The temporal correspondence between the video and audio inputs and the generated listener facial motion is illustrated in Figure \ref{fig:fig3}. 

For the \textbf{clips}, we design a visual encoder to map the raw image sequences into the latent space, while mapping the raw audio sequences into mel-spectrograms. For the encoders, we employ the lightweight convolutional networks ResNet-18 \cite{he2016deep} as the backbone. To preserve spatial information in the video, we replace global average pooling with spatial softmax pooling. Finally, the video and audio are concatenated along the depth dimension in the latent space while maintaining temporal alignment, forming the multimodal input \( M \in R^{b,l,d_{v}+d_{a}} \), where $b$ is the batch size, $d_v$ and $d_a$ are the dimension of video and audio in latent space respectively. It can be described as follows:

\begin{equation}
 M_i=concat(Res(v_{i:(i+1)\times l}),Mel(a_{i:(i+1)\times l}))
  \label{eq:1}
\end{equation}


\subsection{Facial Action Diffusion (FAD)}
\label{sec:fad}

Unlike previous methods, we formulate the task of visual and auditory perception-based listener facial expression generation as a denoising diffusion probabilistic process \cite{ho2020denoising}, often called Stochastic Langevin Dynamics \cite{welling2011bayesian}. 

The generation process is regarded as a denoising process. Starting from $X^K$ sampled from Gaussian noise, $X^K$ undergoes $ K $ iterations of the denoising process. As the noise level gradually decreases, it eventually generates the desired noise-free result $X^0$. The process of 
 $X^{k-1}\sim p_{\theta }(X^{k-1}|X^k)$ can be represented by the following equation:

\begin{equation}
 X^{k-1}=\alpha (X^k-\gamma \varepsilon _{\theta }(X^k,k)+ \mathcal{N}(0, \sigma ^2I)),
  \label{eq:2}
\end{equation}
where $\mathcal{N}(0, \sigma ^2I))$ is the Gaussian noise added at each iteration, $\varepsilon _{\theta }$ is the representation network in the latent space with parameter $\theta$, and $\alpha$,$\gamma$,$\sigma$ as a function of the iteration steps, also called the noise schedule. Equation \ref{eq:2} describes a single iteration in the generation process, following \cite{ho2020denoising}.

To complete the reverse process of the diffusion model, we need to construct and optimize a neural network. Note that we design ELNet as the noise prediction network $\varepsilon _{\theta }$, which will be described in Section \ref{sec:elnet}.
The entire denoising process is defined as a Markov chain with learned Gaussian transitions, starting from $p(X^K)=\mathcal{N}(X^K;0,I) $:

\begin{equation}
p_{\theta }(X_{0:K}):=p(X^K)\prod_{k=1}^{K}p_{\theta }(X^{k-1}|X^k)   \label{eq:3}
\end{equation}

During the inference process, we consider embedding the multimodal speaker inputs as conditional probabilities. On this basis, we regard the \(i\)-th multimodal input \textbf{clip} as a conditional probability distribution. This process can be represented from Equation \ref{eq:2} to the following formula:

\begin{equation}
 F_i^{k-1}=\alpha (F_i^k-\gamma \varepsilon _{\theta }(F_i^k,k, M_i)+ \mathcal{N}(0, \sigma ^2I))
  \label{eq:4}
\end{equation}
Equation \ref{eq:4} describes the iterative steps when we input a single \textbf{clip} as a unit, combined with conditional probabilities.

The training process is the inverse of the noise addition process, which is the denoising process. The training process first acquires the ground-truth listener facial expressions from the dataset. During the \(K\) iterations, appropriate noise $\epsilon ^k$ is progressively added.
The process follows the distribution $X^{k}\sim q(X^{k}|X^{k-1})$. The loss at each iteration is represented by the following equation:

\begin{equation}
\mathcal{L}= MSE(\epsilon ^k, \varepsilon _{\theta }(X^0+\epsilon ^k,k))
  \label{eq:5}
\end{equation}

To generate samples from the given speaker video and audio, we consider the multimodal inputs as conditional probabilities and the loss is modified from Equation \ref{eq:5} to:

\begin{equation}
\mathcal{L}= MSE(\epsilon ^k, \varepsilon _{\theta }(F_i^0+\epsilon ^k,k, M_i))
    \label{eq:6}
\end{equation}
Equation \ref{eq:6} is the sole loss used in our model training, describing the loss at each stage of the training iterations.

\subsection{Efficient Listener Network (ELNet)}
\label{sec:elnet}
The overview of the framework in Figure \ref{fig:fig2} shows the overall architecture of ELNet. It utilizes a U-Net\cite{ronneberger2015u} as the backbone. We have made the following modifications to adapt it to our training framework.

The U-Net architecture is simple, efficient to train, and has achieved success in the field of image generation. To reduce the parameter count and improve the running speed, we apply the network to learn feature representations in the latent space. Unlike image generation, where the feature space and motion space are typically two-dimensional, ours are mapped to one dimension. This not only reduces the number of model parameters but also enhances the model's inference speed. Specifically, we employ 1D convolutions from Janner et al. \cite{janner2022planning} to modify the original network. Additionally, we follow the practice in image diffusion networks by replacing the original BatchNorm \cite{ioffe2015batch} in the U-Net with GroupNorm \cite{wu2018group} for more stable training \cite{huang2024stablemofusion}. 

\section{Experiments}
\label{sec:formatting}


\begin{table*}[t]
\centering
\label{tab:1}
\begin{tabularx}{\textwidth}{@{} ll *{8}{>{\centering\arraybackslash}X} @{}}
\toprule
\multicolumn{2}{l}{\textbf{Testing Dataset}} & \multicolumn{4}{c}{\textbf{L2L}} & \multicolumn{4}{c}{\textbf{ViCo}} \\ 
\cmidrule(lr){3-6} \cmidrule(lr){7-10}
\multicolumn{1}{l}{\multirow{2}{*}{\textbf{Groups}}} & \multicolumn{1}{l}{\multirow{2}{*}{\textbf{Methods}}} & 
\multicolumn{1}{c}{L2$\downarrow$} & \multicolumn{1}{c}{FD$\downarrow$} & \multicolumn{1}{c}{SI} & \multicolumn{1}{c}{TLCC} & 
\multicolumn{1}{c}{L2$\downarrow$} & \multicolumn{1}{c}{FD$\downarrow$} & \multicolumn{1}{c}{SI} & \multicolumn{1}{c}{TLCC} 
\\
& & \multicolumn{1}{c}{} & \multicolumn{1}{c}{($10^2$)} & \multicolumn{1}{c}{} & \multicolumn{1}{c}{
} & 
\multicolumn{1}{c}{} & \multicolumn{1}{c}{($10^2$)} & \multicolumn{1}{c}{} & \multicolumn{1}{c}{
} 
\\
\midrule

\multicolumn{1}{l}{\multirow{5}{*}{Classical}} & 
NN motion   & 24.07 & 5.75  & 1.15 & 0.142 & 41.26 & 17.41 & 0.74 & 0.034 \\
& NN audio    & 34.37 & 11.62 & 1.17 & 0.080 & 49.71 & 24.50 & 0.39 & 0.135 \\
& Random      & 26.99 & 8.25  & 1.19 & 0.175 & 35.58 & 13.35 & 0.72 & 0.269 \\
& Mirror      & 43.93 & 17.59 & 1.32 & 0.335 & 48.28 & 21.26 & 1.24 & 0.268 \\
& Median      & 26.91 & 8.35  & 0.00 & --    & 35.56 & 13.42 & 0.00 & --    \\ 
\specialrule{.4pt}{0pt}{0pt}
\addlinespace[2pt]

\multicolumn{1}{l}{\multirow{3}{*}{Deep}} & 
ViCo-LSTM \cite{zhou2022responsive}   & 36.73 & 13.37 & 1.03 & 0.067 & 74.00 & 54.80 & 0.52 & 0.092 \\
& L2L \cite{ng2022learning}        & 20.30 & 4.67  & 1.07 & 0.072 & 38.37    & 16.12    & 0.65   & 0.002    \\
& ReactFace \cite{luo2024reactface}  & 41.68 & 18.79 & 0.50 & 0.069 & 67.85 & 46.72 & 0.16  & 0.011  \\
\specialrule{.4pt}{0pt}{0pt}
\addlinespace[2pt]
\multicolumn{2}{l}{GT} & -- & -- & 1.17 & 0.198 & -- & -- & 0.72 & 0.255 \\ 
\midrule

\multicolumn{2}{l}{\textbf{Ours}} &\textbf{ 2.35} & \textbf{0.05} & 0.14 & 0.004 & \textbf{2.31} & \textbf{0.05} & 0.98 &  0.009  \\ 
\bottomrule
\end{tabularx}
\caption{ \textbf{Expression Comparison.} Comparisons of the generated facial expressions with ground-truth annotations (GT) on two popular datasets. The $\downarrow$ indicates that lower values are better; the best results are highlighted in bold.}
\end{table*}

\begin{table*}[t]
\centering
\label{tab:2}
\begin{tabularx}{\textwidth}{@{} ll *{8}{>{\centering\arraybackslash}X} @{}}
\toprule
\multicolumn{2}{l}{\textbf{Testing Dataset}} & \multicolumn{4}{c}{\textbf{L2L}} & \multicolumn{4}{c}{\textbf{ViCo}} \\ 
\cmidrule(lr){3-6} \cmidrule(lr){7-10}
\multicolumn{1}{l}{\multirow{2}{*}{\textbf{Groups}}} & \multicolumn{1}{l}{\multirow{2}{*}{\textbf{Methods}}} & 
\multicolumn{1}{c}{L2$\downarrow$} & \multicolumn{1}{c}{FD$\downarrow$} & \multicolumn{1}{c}{SI} & \multicolumn{1}{c}{TLCC} & 
\multicolumn{1}{c}{L2$\downarrow$} & \multicolumn{1}{c}{FD$\downarrow$} & \multicolumn{1}{c}{SI} & \multicolumn{1}{c}{TLCC} \\
& & \multicolumn{1}{c}{} & \multicolumn{1}{c}{($10$)} & \multicolumn{1}{c}{} & \multicolumn{1}{c}{
} & 
\multicolumn{1}{c}{} & \multicolumn{1}{c}{($10$)} & \multicolumn{1}{c}{} & \multicolumn{1}{c}{
} \\
\midrule

\multicolumn{1}{l}{\multirow{5}{*}{Classical}} & 
NN motion   & 6.01  & 5.18  & 0.82  & 0.087 & 10.89 & 13.73 & 0.77 & 0.199 \\
& NN audio    & 9.17  & 10.61 & 0.86  & 0.109 & 11.86 & 15.97 & 0.60 & 0.159 \\
& Random      & 6.92  & 6.22  & 0.71  & 0.044 & 8.68  & 9.53  & 0.38 & 0.001 \\
& Mirror      & 11.23 & 13.43 & 1.02  & 0.543 & 10.63 & 12.55 & 0.95 & 0.448 \\
& Median      & 6.91  & 6.23  & 0.00  & --    & 8.67  & 9.52  & 0.00 & --    \\ 
\specialrule{.4pt}{0pt}{0pt}
\addlinespace[2pt]

\multicolumn{1}{l}{\multirow{3}{*}{Deep}} & 
ViCo-LSTM\cite{zhou2022responsive}   & 9.53  & 1.03  & 0.66  & 0.221 & 15.42 & 2.60  & 0.47 & 0.072 \\
& L2L \cite{ng2022learning}         & 5.56  & 4.01  & 0.78  & 0.094 & 10.08    & 16.12    & 0.65   & 0.002    \\
& ReactFace \cite{luo2024reactface}   & 11.54 & 15.3  & 0.59  & 0.453 & 13.45 & 20.50   & 0.50   & 0.023 \\
\specialrule{.4pt}{0pt}{0pt}
\addlinespace[2pt]

\multicolumn{2}{l}{GT} & --  & -- & 0.84 & 0.088 & --  & -- & 0.66 & 0.057 \\ 
\midrule

\multicolumn{2}{l}{\textbf{Ours}} & \textbf{0.22} & \textbf{0.05} & 1.40 & 0.007 &\textbf{ 0.34} & \textbf{0.01}  & 0.99 &0.008\\ 
\bottomrule
\end{tabularx}
\caption{\textbf{Rotaton Comparison.} Comparisons of the generated jaw rotation and head rotation with ground-truth annotations (GT) on two popular datasets. The $\downarrow$ indicates that lower values are better; the best results are highlighted in bold.}
\end{table*}

\subsection{Experimental Setup}
\paragraph{Datasets.}
We evaluate the effectiveness of our method on two popular conversational datasets, the ViCo dataset \cite{zhou2022responsive} and L2L dataset \cite{ng2022learning}. The ViCo dataset consists of 483 conversational videos collected from the internet, totaling 95 minutes and 22 seconds. Each video includes a pair of speaker and listener videos. To obtain the listener facial motion representation, we employ DECA \cite{feng2021learning} as a tool to detect the 3DMM coefficients for each frame of the listener video.
The L2L dataset contains 72 hours of interview videos collected from six YouTube channels. It includes the listener 3DMM coefficients as well as the timestamps of when the conversations occurred. For the dataset, we follow the L2L dataset by setting T to 64. According to the FLAME 3DMM \cite{li2017learning}, it defines 50 expression coefficients ($d_{exp}=50$), 3D jaw rotation ($d_j=3$), and 3D head rotation ($d_h=3$). For L2L dataset, We follow their approach to split the dataset into 70\% for training, 20\% for validation, and 10\% for testing. For the ViCo dataset, we follow their practice of using 398 videos for training, 42 for validation, and 43 for testing. Additionally, we sample from the ViCo according to the predefined \( T \) to maintain consistency with the format of the L2L dataset.
\paragraph{Implementation Details.} 
To process the video, we set the frame rate to 30 FPS and process the video data on a per-frame basis. The employed ResNet-18 network outputs \( d_v = 192 \) in the spatial softmax pooling layer. To process the audio, we set the sampling rate to 16 kHz to extract the mel-spectrogram ($d_a=128$). We set the \textbf{clip} length l to 8 and resize the video frames to a resolution of 96x96. During the training phase, we set \( K \) to 100 and employ a squared cosine schedule \cite{nichol2021improved}. We use the AdamW optimizer \cite{kingma2014adam} with a learning rate of \( 1e^{-4}\), \( \beta_1 = 0.95 \), and \( \beta_2 = 0.999\) to train the model for 20 epochs on a single Tesla A40 GPU. During the inference phase, we set the number of iterations to 1.
\paragraph{Baselines.}
We categorize the baselines into classical methods based on handcrafted features and end-to-end deep learning methods. The classical methods follows L2L \cite{ng2022learning} and are as follows:

\begin{itemize}[]
\item[\textbullet] \textbf{NN motion/NN audio}: Given the speaker facial motion/audio as input, we search for the nearest neighbor listener facial motion from the training set as the output.
\item[\textbullet] \textbf{Random}: We randomly sample a sequence of \( T \) frames of the listener's motion from the training set.
\item[\textbullet] \textbf{Mirror}: We directly use the speaker motion as the output.
\item[\textbullet]  \textbf{Median}: We take the median expression of the listener's motion from the training set and replicate it for \( T \) frames.
\end{itemize}

For deep learning methods, we select state-of-the-art methods and train and test them on the two datasets. The deep learning methods are as follows:

\begin{itemize}[]
\item[\textbullet] \textbf{ViCo-LSTM \cite{zhou2022responsive}}: For fairness, we use the official code with an LSTM backbone to learn the listener's motion from the speaker facial coefficients extracted from video and audio features.
\item[\textbullet] \textbf{L2L \cite{zhou2022responsive}}: It introduces VQ-VAE \cite{van2017neural} to learn the mapping of 3DMM coefficients to a codebook. The network is designed to model the listener's motion using the speaker facial motion and audio as input.
\item[\textbullet] \textbf{ReactFace \cite{luo2024reactface}}: It designs a network that takes the speaker video and audio as inputs to generate diverse listener facial motions.
\end{itemize}

\paragraph{Metrics.}
We use the following metrics to evaluate the generated expression and rotation coefficients:

\begin{itemize}[]
\item[\textbullet] {L2 Distance }: The L2 distance between the ground truth and the generated expression/rotation.
\item[\textbullet] {Fréchet Distance (FD) \cite{heusel2017gans} }: To assess the realism of the generated expressions, we compute the Fréchet distance between the generated expressions and rotations and their corresponding ground truths.
\item[\textbullet] {SI }: We follow the L2L \cite{ng2022learning} approach by using k-means clustering on all listener expressions and rotations in the training set. We then report the Shannon index of the predicted results. The clustering parameters are kept consistent with those in L2L.
\item[\textbullet] {TLCC }: We analyzed the lagged relationship between the generated listener and the input speaker by calculating the time-lagged cross-correlation (TLCC) \cite{boker2002windowed}.

\end{itemize}

\subsection{Comparison Results}
We retrained L2L \cite{ng2022learning} on the ViCo dataset \cite{zhou2022responsive}. And we retrained ViCo-LSTM \cite{zhou2022responsive} and ReactFace \cite{luo2024reactface} on the ViCo dataset \cite{zhou2022responsive}, as well as on the L2L dataset \cite{ng2022learning}. As can be seen from the results in Tables 1 and 2, our method achieves significant advantages in generating both facial expressions and rotations, as evidenced by the L2 and Fréchet distances (FD). Compared to other end-to-end methods, our proposed paradigm focuses more on establishing the mapping from visual and auditory perception to the high-dimensional motion space. Therefore, prediction accuracy is an important metric for evaluating our method. The experimental results on two datasets fully demonstrate the effectiveness of our proposed new paradigm.
Although the method L2L achieves better results in terms of facial expression diversity on the L2L dataset, our results on the ViCo dataset show that our method is closer to the ground truth (GT) in terms of diversity compared to other methods. With a module specifically designed for diversity, our method can also generate more diverse results in head and jaw rotations. Head movements are often used in communication to express agreement or disagreement, so diverse head generation results play an important role in future human-computer interaction. 

The lag detection results between listener motion and speaker motion indicate that our method can generate results with minimal delay. While high synchronization does not necessarily imply the superiority of a method, the goal of our proposed approach is to rapidly generate listener motions. Therefore, the synchronization metric demonstrates that our method can effectively avoid degradation of interaction quality caused by latency when processing speaker input. In summary, compared to state-of-the-art methods, our proposed approach can effectively generate realistic listener facial motion predictions.

\paragraph{Efficiency of our method.}
To evaluate the computational efficiency of our method, we compared the FLOPs and running time with existing deep learning methods, see Table 3. For models that require the extraction of 3DMM coefficients, we used DECA \cite{feng2021learning} as the pre-extractor and calculated the FLOPs and running time for the entire pipeline. Our testing conditions involved using a speaker video and audio with \( T = 64 \) frames as input, and all models were tested with a batch size of 1. During this process, we only measured the model inference time, excluding the time for data loading and transformation. Models that directly use visual and auditory perception as inputs have a significant computational complexity advantage over those that require pre-extraction of coefficients. The results in the table show that the computational complexity of methods based on extracting coefficients using 3DMM is over 40 times higher than ours, with a running time that is more than 600 times longer. This fully demonstrates the performance bottleneck we have described.

Although ReactFace also avoids the performance bottleneck associated with 3DMM, our method not only reduces FLOPs by 18.5\% but also increases the running speed by 101 times, which means our method reduces the computational time by 99\% compared to state-of-the-art methods. This indicates that our method not only achieves competitive results but also exhibits high computational efficiency, which is of great significance for real-time facial motion generation.

\begin{table}[t]
\centering
\label{tab:3}
\begin{tabularx}{\linewidth}{@{} l >{\centering\arraybackslash}X >{\centering\arraybackslash}X @{}}
\toprule
\textbf{Method}             & \textbf{FLOPs (G)}          & \textbf{Times (s)}          \\
\midrule
ViCo-LSTM \cite{zhou2022responsive}                & $4646.86_{\scriptscriptstyle 47.2\times}$ & $6.82_{\scriptscriptstyle 620\times}$ \\
L2L \cite{ng2022learning}                       & $4659.22_{\scriptscriptstyle 47.3\times}$ & $7.19_{\scriptscriptstyle 653.6\times}$ \\
ReactFace \cite{luo2024reactface}                & $120.77_{\scriptscriptstyle 1.2\times}$   & $1.12_{\scriptscriptstyle 101.8\times}$ \\
\midrule
\textbf{Ours}             & $\mathbf{98.31_{\scriptscriptstyle 1\times}}$ & $\mathbf{0.011_{\scriptscriptstyle 1\times}}$ \\
\bottomrule
\end{tabularx}
\caption{FLOPs and running time comparison of our method with state-of-the-art methods.We boldface the methods with the smallest FLOPs consumption and the fastest runtime.}
\end{table}

\subsection{Ablation Studies}
Effectively perceiving the information conveyed by the speaker is an essential requirement for generating appropriate listener facial actions. Therefore, we design ablation studies to demonstrate the importance of multimodal data for listener facial motion generation. Table 4 quantifies the results using different inputs. 
We show that using both audio and video as multimodal inputs yields better performance compared to using only the speaker audio or video as input. 

\begin{table}[h]
\centering
\label{tab:4}
\begin{tabularx}{\linewidth}{@{} ll *{8}{>{\centering\arraybackslash}X} @{}}
\toprule
\multicolumn{2}{l}{} & \multicolumn{2}{c}{\textbf{Expression}} & \multicolumn{2}{c}{\textbf{Rotation}} \\ 
\cmidrule(lr){3-6} \cmidrule(lr){7-10}
\multicolumn{1}{l}{\multirow{1}{*}{\textbf{audio}}} & \multicolumn{1}{l}{\multirow{1}{*}{\textbf{video}}} & 
\multicolumn{1}{c}{L2$\downarrow$}  & \multicolumn{1}{c}{SI}  & 
\multicolumn{1}{c}{L2$\downarrow$}  & \multicolumn{1}{c}{SI}  \\
\midrule

\multicolumn{1}{c}{{\checkmark}} & 
    & 2.41  & 0.986  & 0.40  & 0.964  \\
\multicolumn{1}{c}{{}} & 
\checkmark   & 2.35  & 0.985 & 0.36  & 0.961\\
\multicolumn{1}{c}{{\checkmark}} & 
\checkmark      & \textbf{2.31}  & 0.989   & \textbf{0.34}  & 0.996  \\

\bottomrule
\end{tabularx}
\caption{Ablation Study Results on ViCo Dataset \cite{zhou2022responsive}. We select L2 and SI as metrics for accuracy and diversity, respectively. The best results are highlighted in bold.}
\end{table}

\subsection{Meta Parameter Studies}
\begin{figure}[t]\centering
    \captionsetup{type=figure}
    \includegraphics[width=\linewidth]{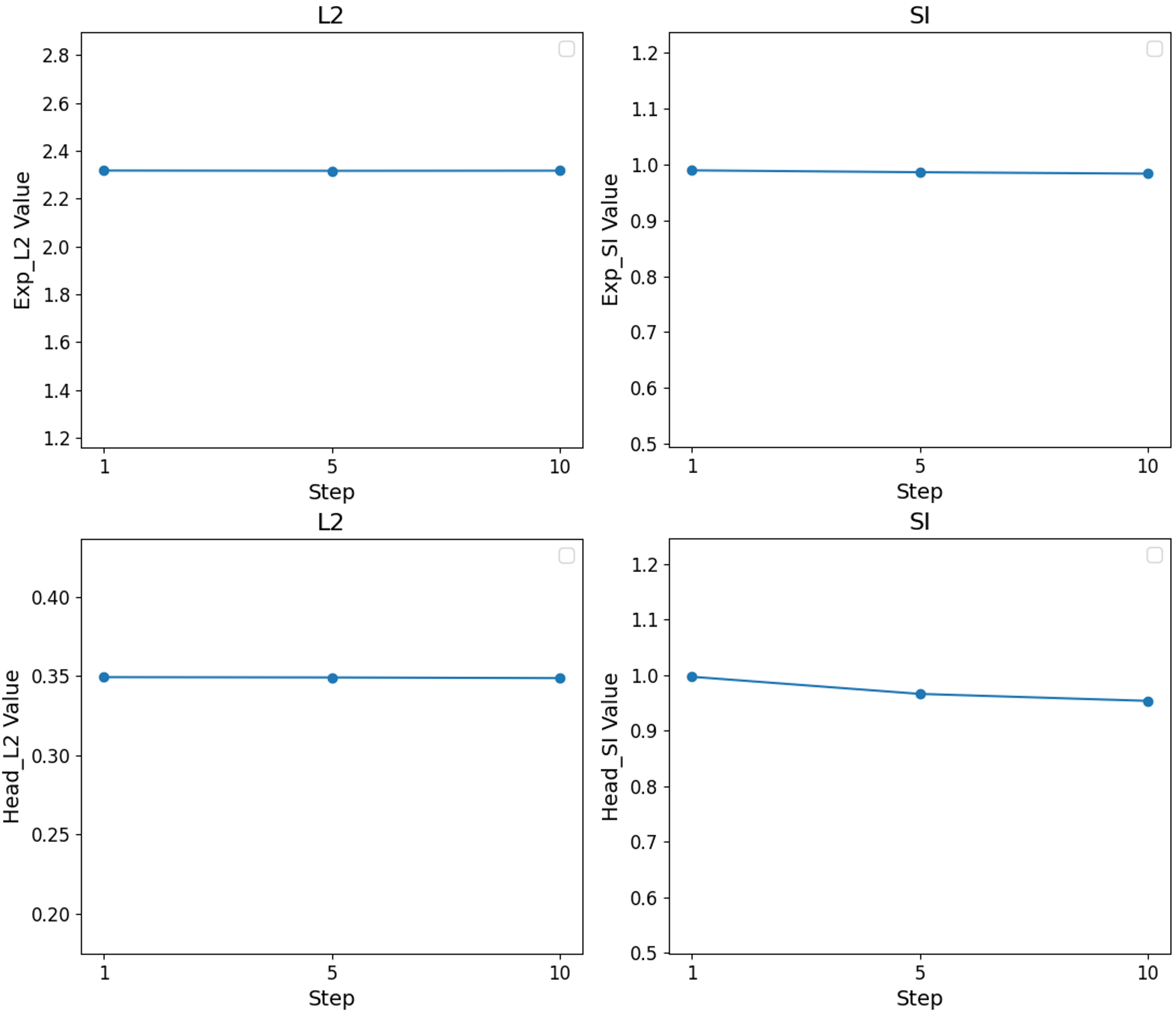}
    \captionof{figure}{The meta parameter studies results present the metrics for different numbers of iterations \( K \) during the inference process. We tested with \( K = 1, 5, 10 \) on the ViCo dataset. We select L2 and SI as metrics for accuracy and diversity, respectively.
    }
  \label{fig:fig4} 
\end{figure}

We discussed the impact of different numbers of iterations during the inference process on the generation performance. 
We assessed the performance variations of the inference results by setting $ K=1,5,10$ on the ViCo dataset and generated corresponding performance curves. 
As shown in Figure \ref{fig:fig4}, despite using a single-step denoising process, the performance of our model is not significantly affected. 
This aligns with intuition, as each iteration in the training phase of diffusion models is not independent but rather learns the mapping relationship between different noise levels and the ground truth \cite{songscore}. Our model learns a prior distribution in the latent space, which corresponds to the mapping at each stage.
This result is particularly important because the time complexity of the denoising process is \( O(K) \); increasing the number of iterations leads to reduced computational efficiency. Our experimental results demonstrate that reducing the number of inference iterations can improve generation efficiency without compromising the quality of the generated results.

\subsection{Qualitative Results}
\begin{figure}[h]\centering
    \captionsetup{type=figure}
    \includegraphics[width=0.7\linewidth]{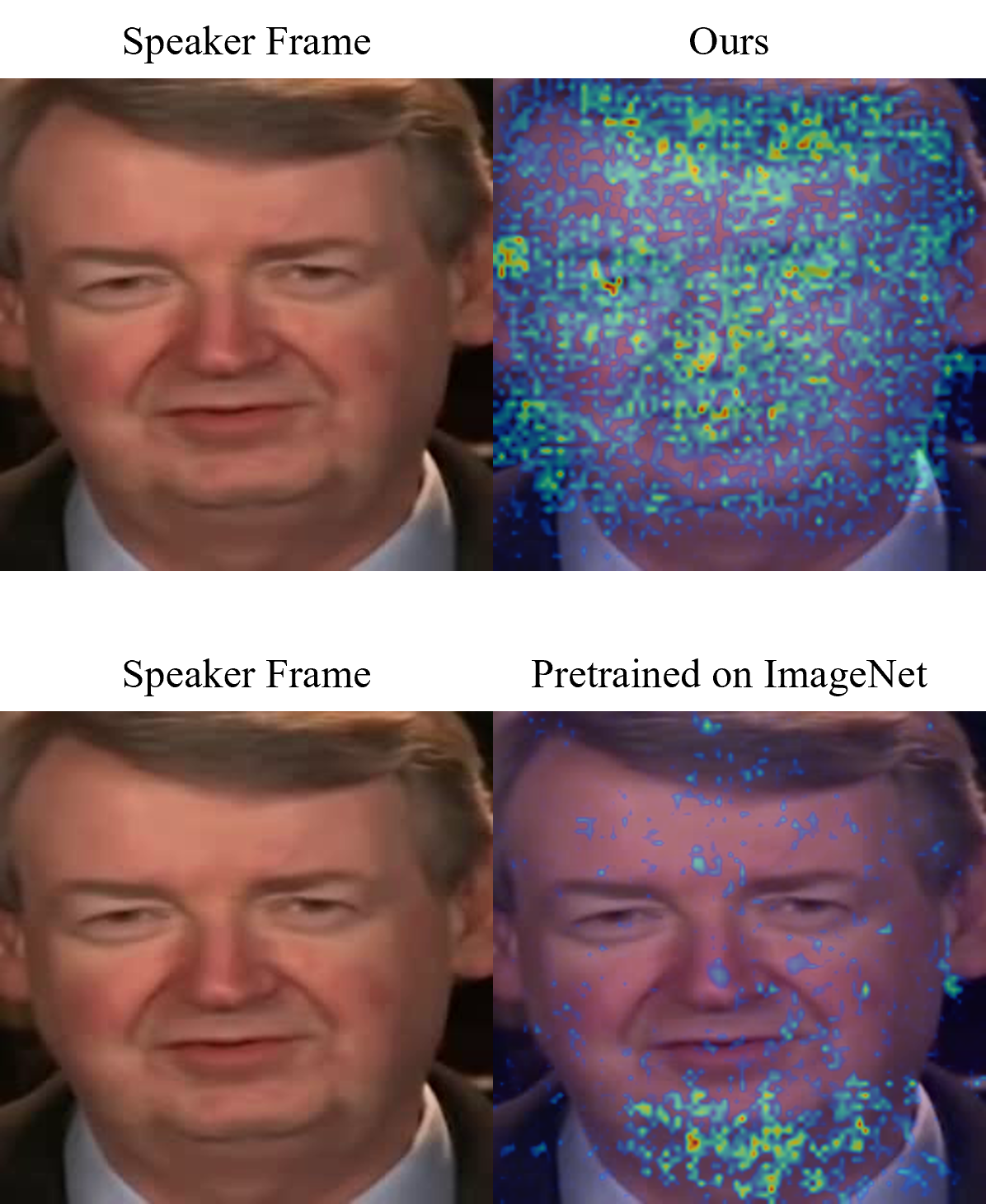}
    \captionof{figure}{We highlight all the contribution feature visualization results. The above shows the comparison results of the ResNet-18 in our framework and the parameters pretrained on ImageNet. We extracted a frame of the speaker from the ViCo dataset \cite{zhou2022responsive} as an example.
    }
  \label{fig:fig5} 
\end{figure}

Our proposed method directly utilizes visual perception. Hence, we focused on the weight distribution of the image feature network. During the inference stage, we input a speaker frame and computed the gradient of the final feature map to the input image via backpropagation, following \cite{selvaraju2017grad}. Here, we averaged the gradients across the channel dimension to obtain the heat map. As shown in Figure \ref{fig:fig5}, we compared our model with the model pretrained on ImageNet. Our trained model can better focus on the speaker face. Attention is concentrated on the facial contours and around the facial features. This indicates that, despite not using 3DMM methods to extract the speaker coefficients, our model is still able to focus on these features.

\section{Conclusion \& Discussion}

In this work, we propose a novel paradigm for modeling listener facial motions. We avoid the computational bottleneck caused by using coefficients extracted by 3DMM. Moreover, unlike previous methods which attempt to align visual and auditory perception with the facial motion at the feature dimension level,  our proposed paradigm represents a novel exploration in end-to-end listener facial motion generation. We regard the facial motion generation process as a "conditional denoising diffusion process." Multimodal data serve as conditional probabilities to guide the generation of facial motions. Such a paradigm provides a novel approach for future real-time human-computer interaction.

Although our method has pushed the performance boundary of listener facial motion under current conditions, there are still some challenges. First, the generalization ability of models still needs to be further enhanced when facing unseen scenarios. For instance, facial expressions and feedback patterns may vary across different linguistic and cultural backgrounds. This requires stronger adaptability and generalization capabilities. Second, verbal feedback is typically expressed quickly through simple words to convey affirmation or negation. Current datasets lack these representations.
We hope to explore richer and more diverse datasets in the future.

{
    \small
    \bibliographystyle{ieeenat_fullname}
    \bibliography{main}
}

\end{document}